\documentclass[10pt,twocolumn,letterpaper]{article}

\usepackage{iccv}
\usepackage{times}
\usepackage{epsfig}
\usepackage{graphicx}
\usepackage{amsmath}
\usepackage{amssymb}
\usepackage[ruled,vlined]{algorithm2e}

\newcommand{\x}{proto}

\usepackage[pagebackref=true,breaklinks=true,letterpaper=true,colorlinks,bookmarks=false]{hyperref}

\iccvfinalcopy 

\def\iccvPaperID{9603} 

\ificcvfinal\fi

\begin{document}

\title{Diamond in the rough: Improving image realism by traversing the GAN latent space}

\author{Jeffrey Wen \qquad Fabian Benitez-Quiroz \qquad Qianli Feng \qquad Aleix Martinez\\
The Ohio State University\\
{\tt\small \{wen.254, benitez-quiroz.1, feng.559, martinez.158\}@osu.edu}

}

\maketitle
\ificcvfinal\thispagestyle{empty}\fi

\begin{abstract}

In just a few years, the photo-realism of  images synthesized by Generative Adversarial Networks (GANs) has gone from somewhat reasonable to almost perfect largely by increasing the complexity of the networks, e.g., adding layers, intermediate latent spaces, style-transfer parameters, etc. This trajectory has led many of the state-of-the-art GANs to be inaccessibly large, disengaging many without large computational resources. Recognizing this, we explore a method for squeezing additional performance from existing, low-complexity GANs. Formally, we present an unsupervised method to find a direction in the latent space that aligns with improved photo-realism. Our approach leaves the network unchanged while enhancing the fidelity of the generated image. We use a simple generator inversion to find the direction in the latent space that results in the smallest change in the image space. Leveraging the learned structure of the latent space, we find moving in this direction corrects many image artifacts and brings the image into greater realism. We verify our findings qualitatively and quantitatively, showing an improvement in Frechet Inception Distance (FID) exists along our trajectory which surpasses the original GAN and other approaches including a supervised method. We expand further and provide an optimization method to automatically select latent vectors along the path that balance the variation and realism of samples. We apply our method to several diverse datasets and three architectures of varying complexity to illustrate the generalizability of our approach. By expanding the utility of low-complexity and existing networks, we hope to encourage the democratization of GANs. \footnote{Code available at https://github.com/jwen307/diamondintherough}

\end{abstract}


\begin{figure}[t]

\begin{center}
   \includegraphics[width=1.0\linewidth]{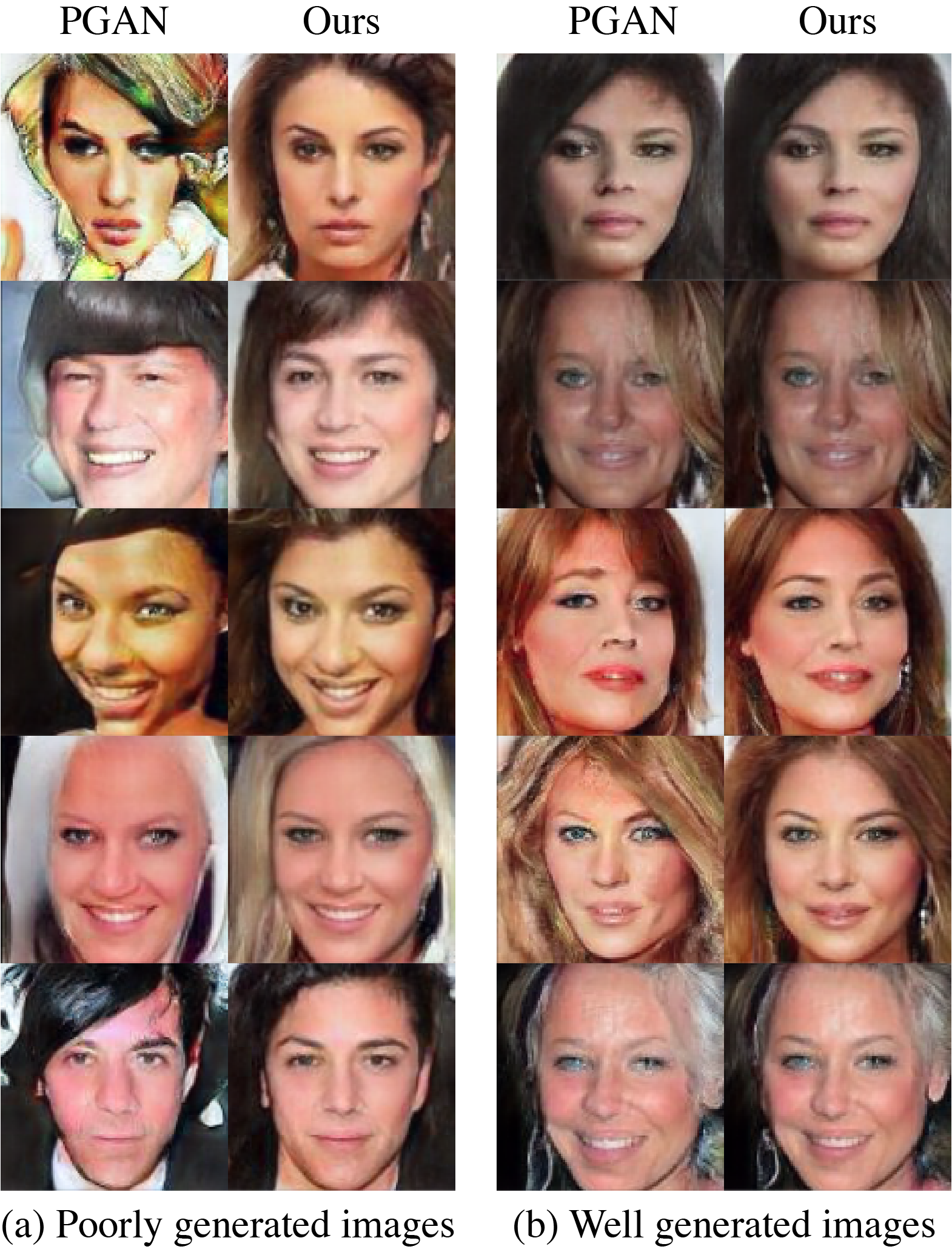}
\end{center}
   \caption{Randomly generated samples from a PGAN trained on CelebA \cite{CelebA} alongside the outputs of our algorithm. The majority of the random images are easily detectable as synthetic. Compare these with the results after applying our proposed framework. Good images generated by PGAN are improved but poorly generated images are dramatically enhanced.}
\label{fig:RandomImages}
\end{figure}

\section{Introduction}


Generative Adversarial Networks (GANs) are powerful algorithms capable of mapping latent vectors ${\bf z}_0$ into novel, photo-realistic images $X$ \cite{KarrasLAHLA20,KarrasLAHLA19,BigGAN,PGAN,Qi:2019a}. Unfortunately, not all latent vectors yield photo-realistic images. Imagine we had a simple algorithm that for every latent vector ${\bf z}_0$ it gave you an alternative latent vector $\hat{\bf z}$ that yielded a much more photo-realistic image $\hat{X}$ of the same class identity and attributes as $X$. This paper defines such an algorithm, Fig. \ref{fig:RandomImages}. We achieve this by constraining the latent space in a specific way that lends to improvement in realism. 

Since the original algorithm derived by \cite{goodfellow}, many variants have been proposed to improve the photo-realism of the generated images \cite{KarrasLAHLA20,KarrasLAHLA19,BigGAN,Karnewar:2020a,PGAN}.  Typically, improvements in photo-realism have come at the cost of increasing the network complexity \cite{KarrasLAHLA19,Donahue:2019a,Razavi:2019a,KarrasLAHLA20}. However, the large computational requirements of these state-of-the-art networks has instilled a disconnect within the GAN community, alienating those without industrial resources and limiting the applicability of GANs for large-scale use. In order to promote the integration of GANs into real-world applications, we must look for accessible, alternative methods.

Here, we take a different approach. Rather than increasing the network's complexity, we constrain the typically used Normal distribution in latent space to a set of great circles on it. We show that these great circles include average sample (\x) images of high realism, Fig. \ref{fig:VisualOverview}. Hence, moving about the great circle toward the converged face and away from the original latent vector improves the realism. We also show that this great circle is easily defined by the geodesic of ${\bf z}_0$ and ${\bf z}_1$, where ${\bf z}_1$ is the estimated latent vector of the image $X$ computed by inverting the GAN's generator function, Fig. \ref{fig:VisualOverview}. Finally, we derive a simple, efficient algorithm to identify a photo-realistic generator latent vector ${\bf \hat{z}}$ about these great circles. A sample of the resulting improved (optimized) images were shown in Fig. \ref{fig:RandomImages}.

We show how the proposed algorithm allows us to achieve consistent photo-realistic images even when using lower-cost networks like PGAN. Through utilizing the full capability of lower-complexity and existing networks, we hope to stimulate the accessibility of GANs.


\section{Related Work}

\subsection{Sampling from areas of high probability}

Brock \etal \cite{BigGAN} proposed to draw latent vectors from a truncated distribution of the original prior $p({\bf z})$. Given a sample drawn from $p({\bf z})$, the values of vector ${\bf z}$ above a given threshold are resampled to be within the threshold. As the threshold becomes smaller, the samples are drawn from a smaller area of higher likelihood resulting in an increase in photo-realism. 

Alternatively, Menon \etal \cite{pulse} approximates the natural image manifold by projecting sampled vectors onto the hypersphere of radius $\sqrt{n}$, $n$ the number of dimensions of the latent space. High-dimensional Gaussian distributions take the form of a ``soap bubble'' with probability mass lying at or near  $\sqrt{n}S^{n-1}$, where $S^{n-1}$ is the $n$-dimensional unit hypersphere. This allows the authors to constrain the latent space in a way that results in increased consistency in the realism of the generated images. 

While these two approaches have shown improvements for specific GAN architectures, they do not fully describe the natural image manifold. As most of the mass of the distribution is on the surface of $\sqrt{n}S^{n-1}$, the truncation trick is not guaranteed to find an area of high probability. Importantly, for architectures like PGAN where the input vector is scaled to be on the hypersphere before passing it to the generator, the approach from \cite{pulse} cannot identify the regions associated to photo-realistic images. 

\begin{figure}[t]
\begin{center}
   \includegraphics[width=1.0\linewidth]{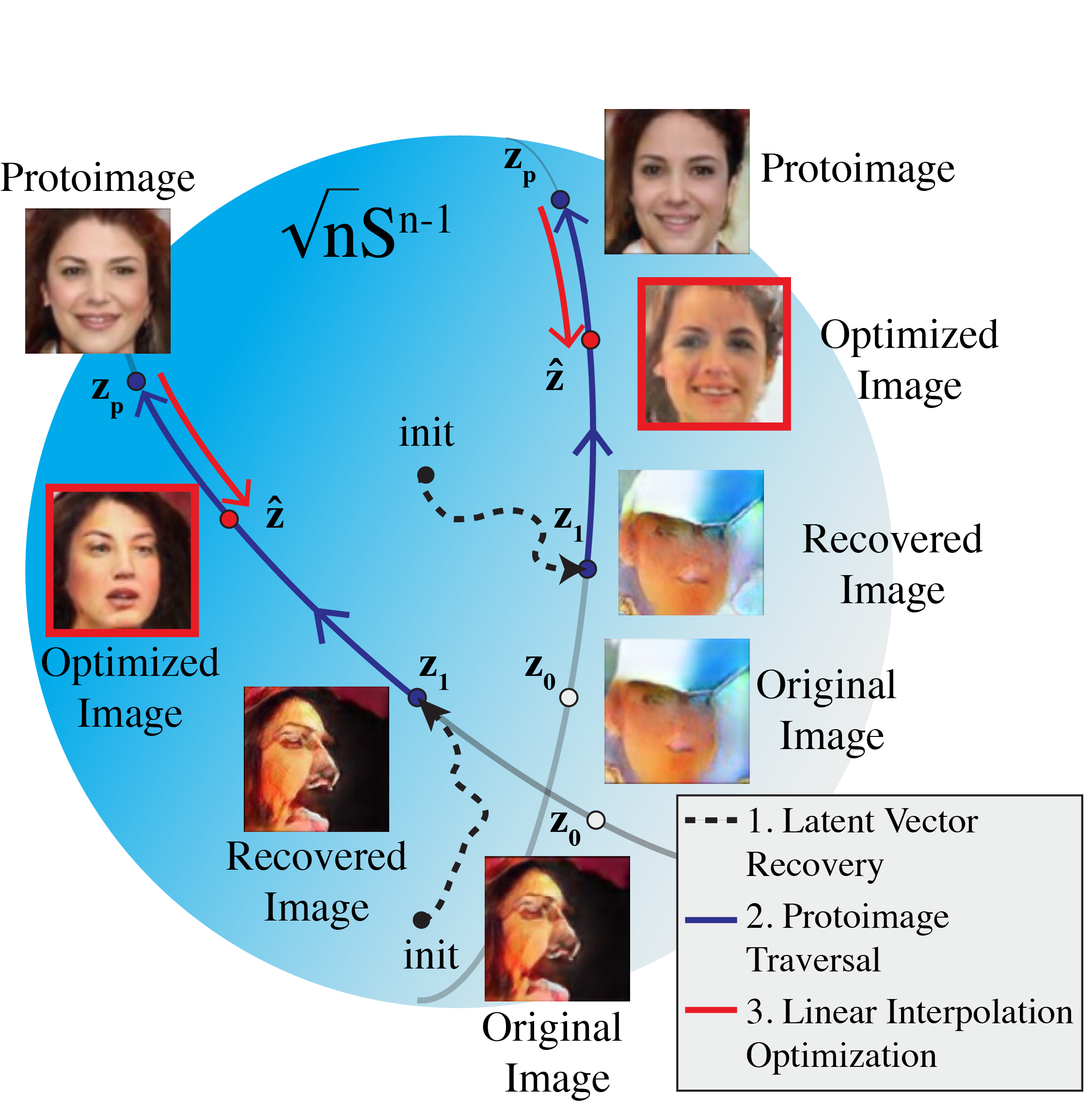}
\end{center}
   \caption{The latent vector ${\bf z}_0$ generates image $X$. Inverting the process we estimate the latent vector ${\bf z}_1$ associated with $X$. The geodesic between ${\bf z}_0$ and ${\bf z}_1$ defines a great circle on the hypersphere. These great circles include a number of highly realistic \x images. Moving towards these \x images about this great circle, increases the realism of the generated image. Our goal is to define a criterion that when optimized yields a photo-realistic image of the same class identity and attributes as $X$.}
\label{fig:VisualOverview}
\end{figure}

Contrary to these approaches, our framework shows that the great circle on this hypersphere in the latent space given by ${\bf z}_0$ and ${\bf z}_1$ contains the photo-realistic images we desire, Fig. \ref{fig:VisualOverview}. This is because this great circle contains a ``\x image'', a highly-realistic sample average of sample images. Hence, moving away from ${\bf z}_0$ and ${\bf z}_1$ increases the realism of the generated image, Fig. \ref{fig:VisualOverview}. 

\subsection{GAN Manipulation} \label{section: ganmanipulation}
Several other recent works have investigated the capability of controlling the behavior of GANs through exploring the latent space. The goal of these methods is to find directions in the latent space that allow for interpretable semantic changes in the resulting images. \cite{radford} discovered the vector arithmetic effect allowing attributes to be added to images by adding an appropriate vector in the latent space. \cite{yang2019semantic,shen2020interpreting} expand this idea by finding a hyperplane where latent vectors on one side exhibit a binary attribute, e.g. presence of glasses, and vectors on the other side do not. By moving latent vectors in the normal direction of the hyperplanes, they are able to semantically manipulate the resulting image for desired attributes. \cite{ganalyze} uses a pretrained assessor network to study the direction of varying image memorability. These methods rely on pretrained classifier networks or manual annotations which may be unavailable or expensive to obtain.

\cite{gansteerability,plumerault20iclr} utilize a self-supervision framework that transforms a generated image in the pixel space and finds the resulting direction of change in the latent space. However, both these approaches are limited to simple transformations such as rotation, zoom, and translation. \cite{Voynov2020UnsupervisedDO} takes an unsupervised approach with a trainable reconstructor network and direction matrix which find directions that are easily distinguishable, and \cite{harkonen2020ganspace,shen2021closedform} learn interpretable directions by performing principle component analysis (PCA) on samples in the latent space or on the learned weights that map the latent vector to the first convolutional layers.

While these methods are able to find directions that allow users to manipulate semantic features like rotation, hair color, lighting, etc., only \cite{shen2020interpreting} discovers a direction that allows for the control of photo-realism. The approach in \cite{shen2020interpreting}, however, requires sampling images and manual annotations of ``good" and ``bad" synthesized images which can be expensive and noisy. Contrary to this approach, our method finds the mentioned direction without supervision by exploiting the learned semantic structure of the latent space to find the direction of the protoimages vectors that represent photo-realistic sample means with well-structured data.

\begin{figure*}[t!]
\begin{center}
   \includegraphics[width=1.0\linewidth]{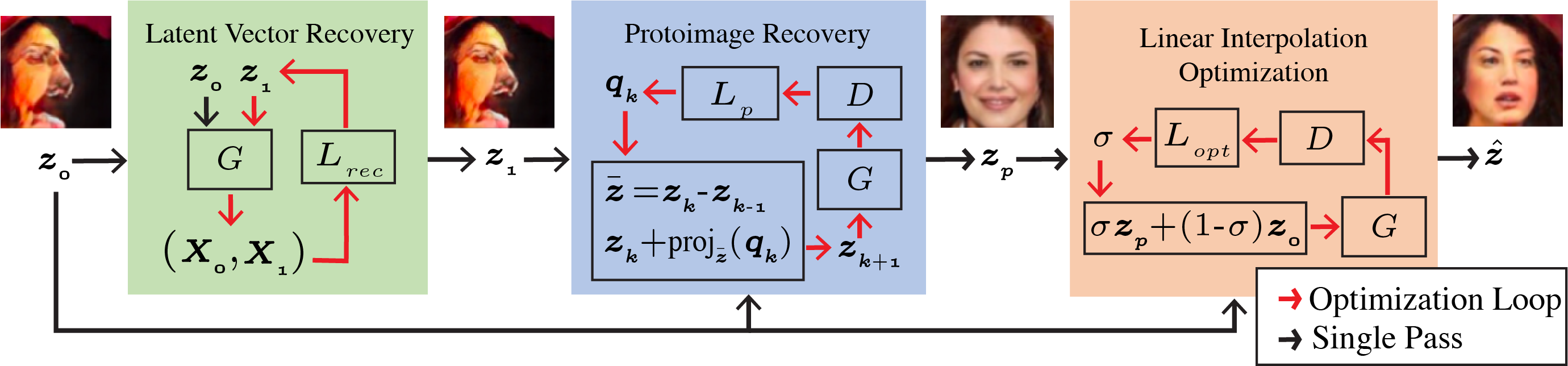}
\end{center}
   \caption{Overview of our framework. We find a latent vector ${\bf z}_1$ that visually resembles the latent vector ${\bf z}_0$ by using the optimization  proposed in \eqref{eq:ObjectiveFuncRecovery} (green box). We traverse the latent space along the great circle that passes through ${\bf z}_0$ and ${\bf z}_1$ to find ${\bf z}_p$ by iteratively projecting the direction of travel onto the hypersphere as explained in \eqref{eq:Mean Image Objective} (blue box). We then use the interpolation technique in \eqref{eq:LinearInterp} to get an optimized image (orange box)}
\label{fig:BigPicture}
\end{figure*}


\subsection{Generator Inversion}
To begin our analysis of the generator latent space, we need a mapping from the image space to the generator's latent space. The inversion of the generator has two common approaches. The first approach is to use an optimization method to minimize the reconstruction loss \cite{Inversion1}\cite{Inversion2}\cite{Inversion3}. Given an initial latent vector ${\bf z} \in \mathbb{R}^n$, the objective function is 
\begin{equation}
\min_z \| X - G({\bf z}) \|_F,
\label{eq:ReconstructionLoss}
\end{equation}
where $X$ is the image, $G({\bf z})$ is the generated image given latent vector ${\bf z}$, and $\| \cdot \|_F$ denotes the Frobenius norm. 

It is also common for a perceptual loss term to be added to this optimization \cite{Johnson:2016,Dosovitskiy:2016a}. The perceptual loss term forces the distance between the original image and recovered image to be small in the feature space of a trained feature extraction network such as a VGG-16 pretrained on ImageNet \cite{VGG16}. We will use both of these criteria in our approach.

Another common approach is to train a separate encoder network \cite{encoder1,encoder2, zhu2020indomain}. Using randomly generated latent vectors and their corresponding generated images, an encoder network can be trained to map the images to the latent vector space. While this method is not sensitive to the initialization of ${\bf z}$ as in the optimization approach, it does typically result in overfitting to the training data. For these reasons, we elected to use the optimization approach together with a perceptual loss in our framework, but we show that one can also use the encoder approach efficiently in our framework.


\section{Methods}

Let the latent vector ${\bf z_{0}}$ generate image $X$ through generator $G(\cdot)$, \ie $X= G({\bf z_{0}})$. Our framework seeks to find a $\hat{\bf z}$ and corresponding $\hat{X} = G(\hat{\bf z})$ which contains the same semantic attributes as $X$ while appearing more photo-realistic than $X$. 

We achieve this by leveraging the discovery of a subset of similar, high-fidelity images that we call \x images. The latent vector, ${\bf z}_p$, corresponding to a \x image lies along a great circle passing through ${\bf z}_0$ and ${\bf z}_1$. ${\bf z}_1$ is found by inverting the generator to approximate the latent vector which produced $X$. By optimizing the interpolation between ${\bf z}_0$ and ${\bf z}_p$, we are able to find a $\hat{\bf z}$, which satisfies our conditions. 

The overall structure of the framework can be found in Fig. \ref{fig:BigPicture}. A visualization of the movement along the hypersphere was shown in Fig. \ref{fig:VisualOverview}. We detail each of the steps in the next few sections.

\subsection{Latent Vector Recovery} \label{section: recovery}
For a given latent vector, ${\bf z}_{0} \sim N({\bf 0},{\bf I}_n)$, we pass the vector through the generator to obtain the original generated image, $X \in \mathbb{R}^d$, $d$  the number of pixels in the image. Given $X$, the goal is to now find a vector ${\bf z}_1$ which will produce a similar image $G({\bf z}_1)$ to $X$. For this optimization problem, we utilize three loss terms.

1. A reconstruction loss, given by 
\begin{equation}
L_{reconst} = \| X - G({\bf z}_1) \|_{1},
\label{eq:ReconstLoss}
\end{equation}
where $\| \cdot \|_1$ is the L1 norm. We use the L1 norm to prevent outlier-valued pixels from overtaking the optimization.

2. A perceptual loss with the features extracted from the feature space of the discriminator. 
As \cite{radford} demonstrated the utility of the discriminator as a feature extractor for classification, we use the discriminator for our perceptual loss to keep the framework isolated from other networks.  
With this in mind, the perceptual loss term was formatted as
\begin{equation} 
L_{percep} = \| D_{f}(X) - D_{f}(G({\bf z}_1)) \|_{1},
\label{eq:percepLoss}
\end{equation}
where $D_{f}(\cdot)$ are the features from the discriminator.

3. As previously mentioned, the majority of the probability mass of a high-dimensional Gaussian lies on the surface of a $\sqrt{d}S^{n-1}$ hypersphere. A proof of this can be found in the Supplemental Material. To keep our search in the space of high probability, we add a simple L2 regularization term to keep the latent vector search near the surface of the hypersphere. Formally, 
\begin{equation}
L_{reg} =\big| \|{\bf z}_1\|_{2} - \sqrt{n} \big |,
\label{eq:regularizationLoss}
\end{equation}
where $n$ is the dimensionality of the latent space and $\| \cdot \|_{2}$ is the L2-norm. 

Putting all three of these terms together, we arrive at the final objective function,
\begin{equation}
{\bf z}_1* = \arg \min_{{\bf z}_1} L_{reconst} + \alpha L_{percep} + \beta L_{reg},
\label{eq:ObjectiveFuncRecovery}
\end{equation}
where $\alpha$ and $\beta$ are chosen weight parameters. 

\begin{figure}[t]
\begin{center}
   \includegraphics[width=1.0\linewidth]{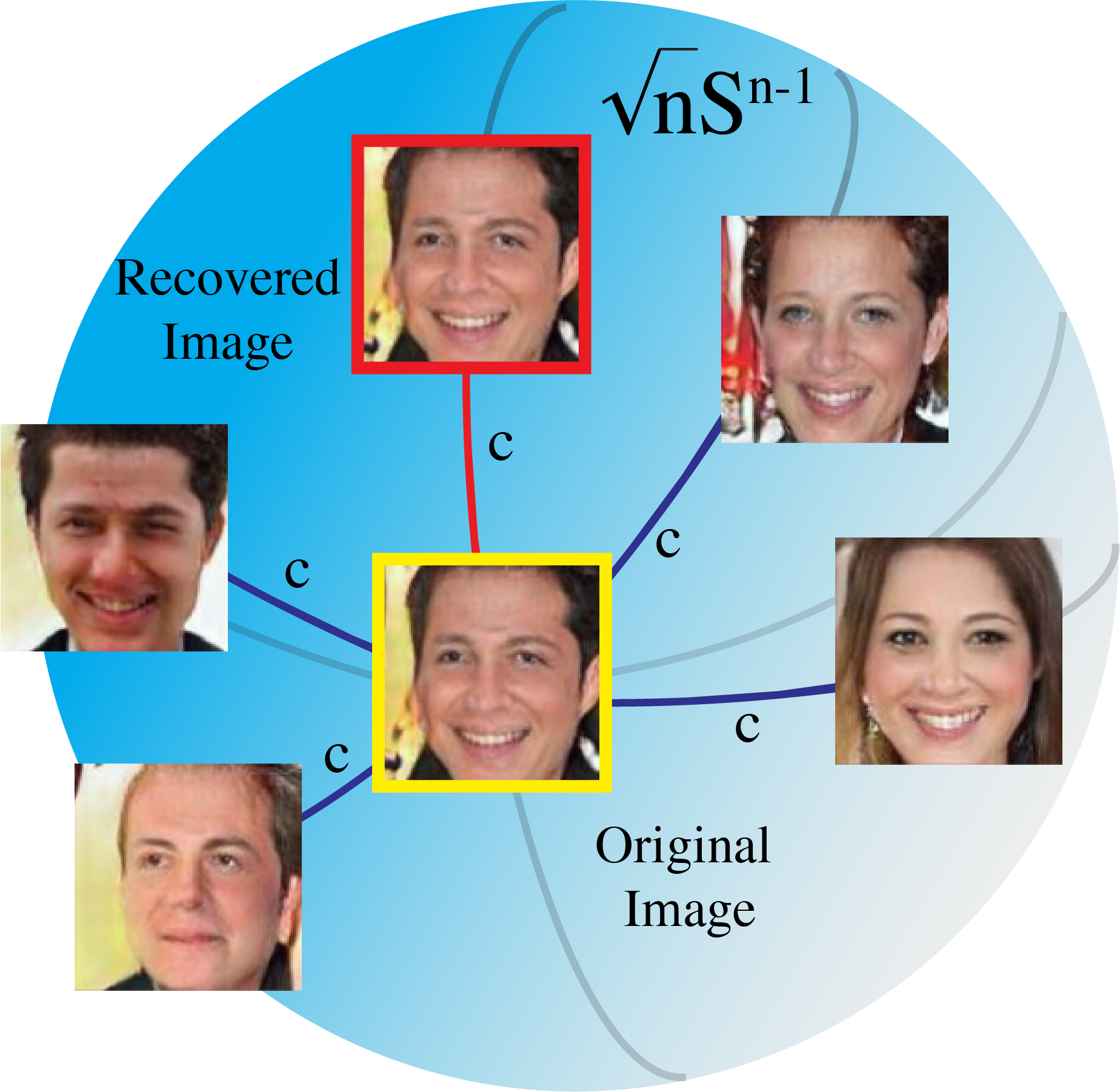}
\end{center}
    \caption{We demonstrate the uniqueness of the latent vectors recovered by our optimization. The recovered image in red is nearly identical to the original image in yellow. Images of an equivalent distance, c, in other directions yield very different images.}
\label{fig: images in random direction}
\end{figure}

The results  of this algorithm can be seen in Fig. \ref{fig:RandomImages}. We note that while the original and recovered images are nearly identical, the pairwise Euclidean distance between the original and recovered latent vectors may not be small. For example, with PGAN on CelebA, the average distance is 25. 

To demonstrate the difference that such a Euclidean distance can make in the pixel-space, we added noise vectors of an equivalent magnitude to the original latent vectors and generated the images of the resulting vectors. These can be seen in Fig. \ref{fig: images in random direction}. While an equivalent distance away, these vectors provide dramatically different images from the original. Thus, this suggests that our recovered latent vectors are in a particular direction where the image attributes do not change very much. We further explore this hypothesis in the next section. 

\subsection{Traversing the Latent Space}\label{walkinlatentspace}
In this section, we explore the effects of moving along the hypersphere in the direction of the recovered latent vector. 

As a simple experiment, we take the difference between the recovered, ${\bf z}_1$, and original latent vector, ${\bf z}_0$ and add it to the recovered latent vector. The resulting vector is projected onto the hypersphere and named ${\bf z}_2$. By repeating the sequence, ${\bf z}_{k+1} = proj({\bf z}_{k} + ({\bf z}_{k}-{\bf z}_{k-1}))$ (where $proj(\dot)$ is the projection onto the hypersphere), we are able to traverse the hypersphere in the direction of the recovered latent vector. The results of this visualization can be seen in Fig. \ref{fig:TraverseImages}.

Note that as the latent vector is moved about the hypersphere, the image quality improves by bringing clarity to the already well-defined faces or by bringing previously unidentifiable faces to images that resemble a face. The key to this process is to note that this walk arrives at a similar set of \x image around the same iteration of the sequence.  These \x images are in fact sample mean images. For the CelebA dataset, for example, the \x image resembles a mean face of a subset of the training set. 

Up until this point, each iteration improves quality but retains most of the unique attributes of the original image. After the \x image, the subsequent iterations provide a completely new identity, which worsens in quality as the walk continues. This behavior is similar to the effects seen using the truncation trick \cite{BigGAN}. \cite{KarrasLAHLA19} report a similar effect with the truncation trick applied to the intermediate latent space. In their experiment, they also show different identities, often antifaces, once the mean image is crossed. However, separate from those experiments, our latent vectors are not converging to the same latent vector. In fact, the latent vectors of the mean images are separated by a large angle. While the effects of moving towards the mean images appear to be the same as the truncation trick, here the generator has designed the latent space to have mean images about every great circle on the hypersphere. 

\begin{figure}[t]
\begin{center}
   \includegraphics[width=1.0\linewidth]{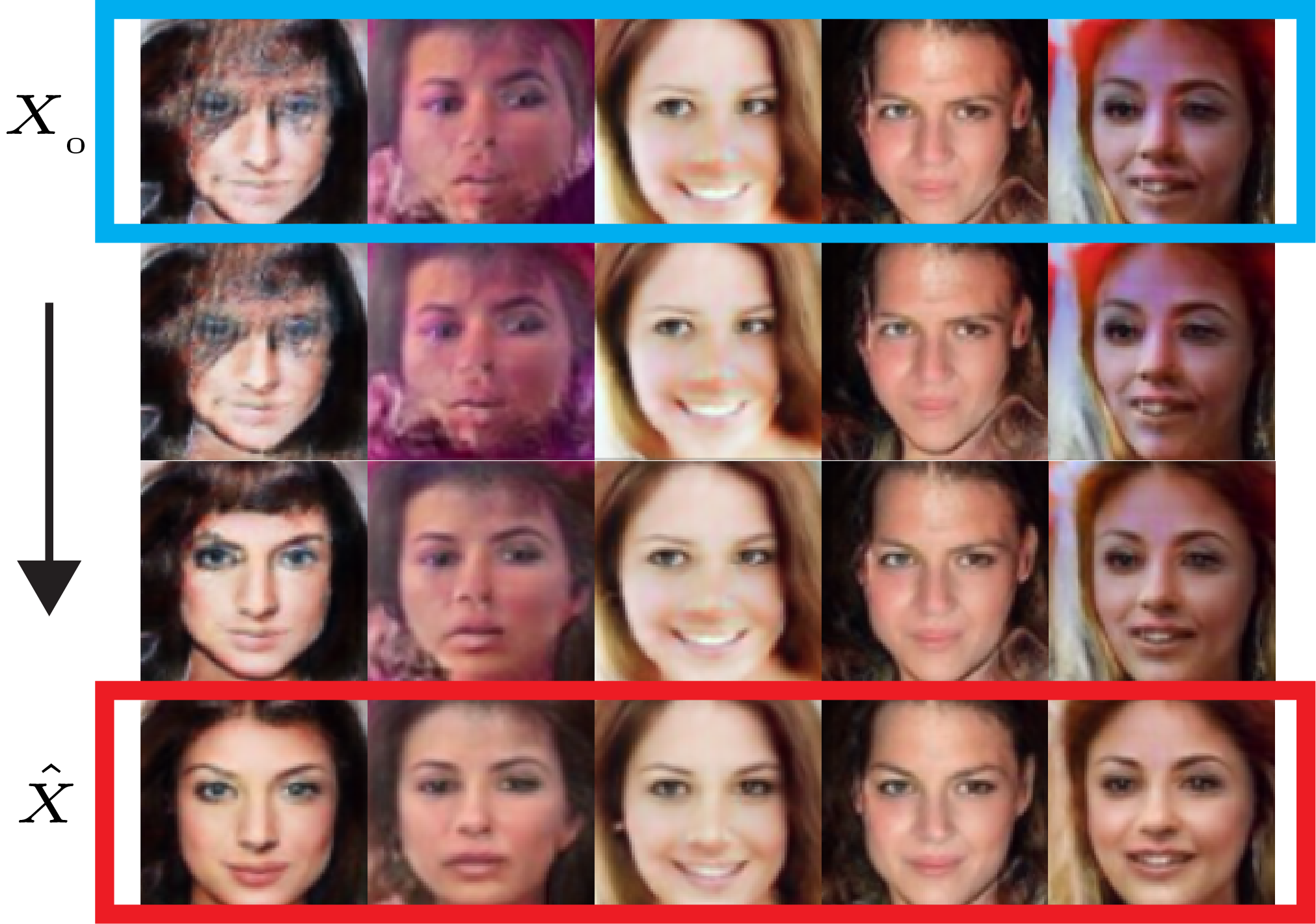}
\end{center}
   \caption{Here we show the progression as we traverse the latent space towards the \x image. The image quality improves along the traversal while maintaining the same attributes as the original image.}
\label{fig:TraverseImages}
\end{figure}

\subsection{Protoimage Recovery}\label{section:protoimage}
As we observed the improvement in image quality near the \x images, we would like to be able to find the locations of these \x images given a random set of latent vectors. 

An interesting feature of some GANs is the inclusion of a minibatch standard deviation layer in the discriminator. This layer encourages variation among the images of an input batch, which results in a lower discriminator score for images that are similar. As the mean images all have a similar look, the resulting discriminator score is very low for a batch of \x images. There is also always a large angle between the original latent vector and the \x image latent vectors. With this in mind, we can establish a criterion for finding the \x images about the hypersphere walk. The objective function below seeks to minimize the discriminator score for a batch of images while also encouraging a small cosine similarity value. Formally,
\begin{equation}
{\bf z}_{p}* = \arg \min_{{\bf z}_{p}} D(G({\bf z}_{p})) + \lambda M({\bf z}_{p},{\bf z}_{0}),
\label{eq:Mean Image Objective}
\end{equation}
where $D(\cdot)$ is the discriminator score, ${\bf z}_{p}$ is the latent vector of the \x image, ${\bf z}_{0}$ is the original latent vector, $\lambda$ is a chosen weight parameter, and $M(\cdot,\cdot)$ is the cosine similarity, which can be expressed as 
\begin{equation}
M({\bf z}_{p},{\bf z}_{0}) = \frac{{\bf z}_{p} \cdot {\bf z}_{0}}{ \| {\bf z}_{p} \| \|{\bf z}_{0} \|}.
\label{eq:CosSim}
\end{equation}

With this objective function, we would like to restrict the search space to vectors along the walk around the hypersphere in the direction of the difference vector. 

First, we initialize a vector ${\bf q}_1$ to control the magnitude and forward/backward direction of the next step. We utilize a vector for controlling the hypersphere traversal to provide more robustness in the optimization. For each iteration, we find the difference vector between the current latent vector and the previous. For the first iteration, this is the difference between the recovered ${\bf z}_{1}$ and original latent vector, ${\bf z}_{0}$. The vector, ${\bf q}_{k}$, is projected onto the difference vector and added to the previous vector. The new vector, ${\bf z}_{k+1}$, is, then, projected onto the hypersphere. The loss is calculated using \eqref{eq:Mean Image Objective}, and this loss is backpropagated through the network to update ${\bf q}_{k+1}$. The entire algorithm is summarized in Algorithm \ref{Mean Image Algorithm}. 

\begin{algorithm}
\SetAlgoLined 
 $\mathbf{z}_{0} \in \mathbb{R}^{n}$ (original latent vectors) \\
 $\mathbf{z}_{1} \in \mathbb{R}^{n}$ (recovered latent vectors) \\
 $\mathbf{q}_{k} \in \mathbb{R}^{n}$ \\
 initialize $\mathbf{q_{1}}$\;
 $k = 1$\;
 $\bar{\mathbf{z}} = \mathbf{z}_{1} - \mathbf{z}_{0}$\;
 $\epsilon  \in \mathbb{R}, \epsilon > 0$ (stopping threshold) \\
 \While{$\| \mathbf{\bar{z}} \|_{2}> \epsilon$ }{
  $\mathbf{z}_{k+1} = \mathbf{z}_{k} + (\mathbf{q}_k^{T} \bar{\mathbf{z}})\frac{\bar{ \mathbf{z}}}{\|\bar{ \mathbf{z}}\|_{2}}$\;
  $\mathbf{z}_{k+1} = \mathbf{z}_{k+1} \frac{\sqrt{n}}{\|\mathbf{z}_{k+1}\|_2}$;
  $L_{p} = D(G(\mathbf{z}_{k+1})) + \lambda \frac{\mathbf{z}_{k+1}^{T}\mathbf{z}_{0}}{\| \mathbf{z}_{k+1} \|_2 \|\mathbf{z}_{0} \|_2}$\;
  Backpropagate to update $\mathbf{q}_{k+1}$\;
  $k=k+1$ \;
  $\bar{\mathbf{z}} = \mathbf{z}_{k} - \mathbf{z}_{k-1}$ \;
 }
 \caption{Finding the \x images}
 \label{Mean Image Algorithm}
\end{algorithm}

\subsection{Optimizing for Improved Images}

With the \x images found, improving any randomly generated image is simply a matter of pushing the latent vectors towards the latent vectors corresponding to the \x images. 

This can be accomplished with an easy linear interpolation between the original latent vector ${\bf z}_0$ and the \x image latent vector, ${\bf z}_{p}$ and then projecting the point onto the hypersphere.
\begin{equation}
\hat{{\bf z}} = proj(\sigma {\bf z}_{p} + (1 - \sigma) {\bf z}_0),
\label{eq:LinearInterp}
\end{equation}
where the value of $\sigma$ controls the proximity to the original latent vector and the \x image latent vector. 

\begin{figure}[t]
\begin{center}
   \includegraphics[width=1.0\linewidth]{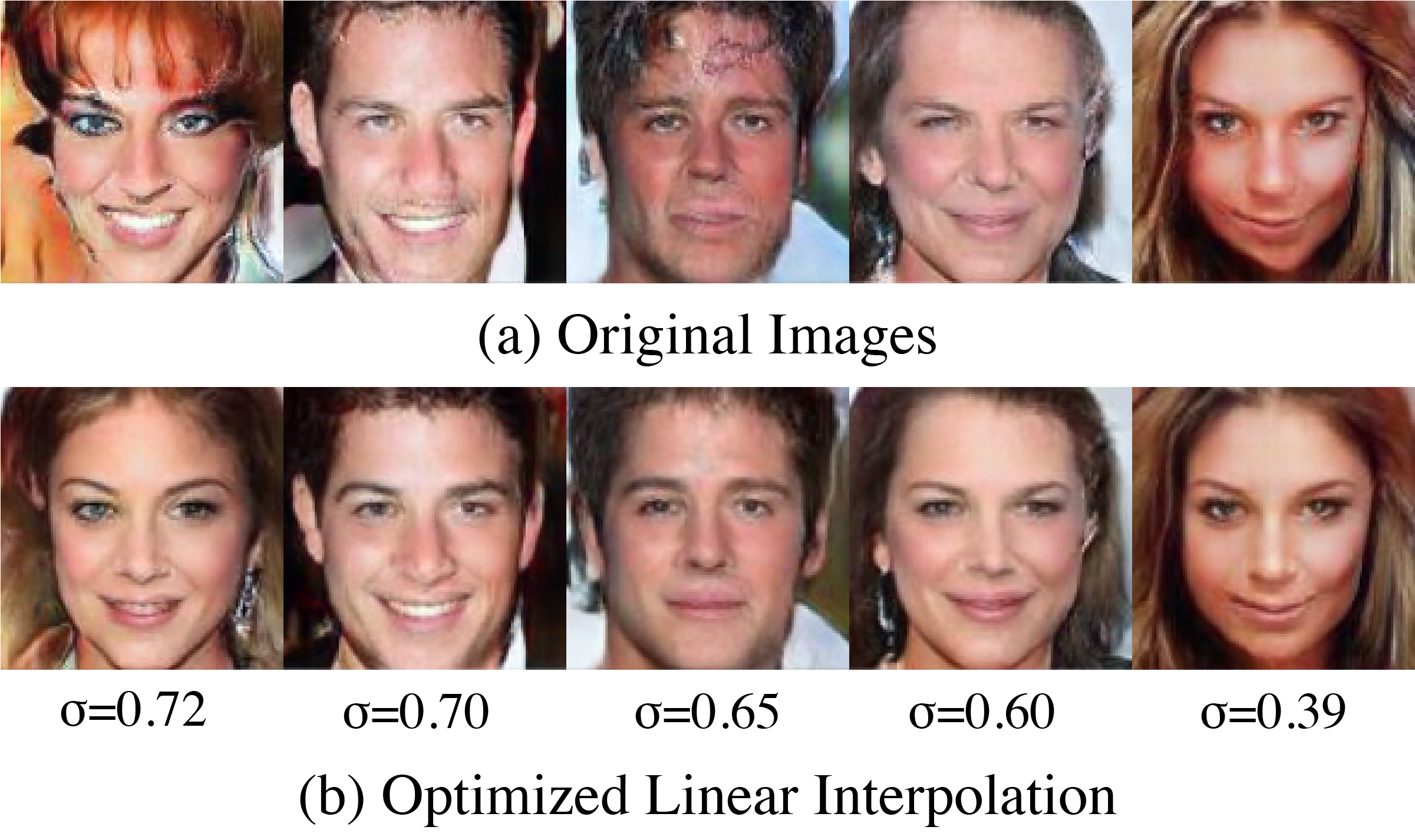}
\end{center}

   \caption{Results of applying our optimization framework to randomly generated images. }
\label{fig:Optimized Images}
\end{figure}

While setting a constant $\sigma$ can yield a decent balance between fidelity improvement and variation as seen in Fig. \ref{fig:Optimized Images}, we recognize that certain images need more improvement and some minor improvements to already good images are not worth the reduction in variation. 

To automate the balance between fidelity and variation, we optimize for the linear interpolation coefficient, $\sigma$. The discriminator is trained to discriminate between real photos and generated images. With the Wasserstein loss \cite{WGAN}, the discriminator acts as a critic, providing higher scores for real images and lower scores for images that are believed to be fake. As the image quality of generated images improves, the discriminator score improves. However, as the improved images begin to lose variation, the discriminator score drops again due to the minibatch standard deviation layer. While one option would be to find the $\sigma$s which provided the highest discriminator score for a batch of images, the optimization tends to favor the extremes, bringing some values of $\sigma$ up very high towards $1$ but dropping $\sigma$ for other images to counteract for the loss of variation. 

To derive a criterion that is beneficial for most images, we instead bring the discriminator score to the middle ground between scores given for real images and scores given for poorly generated images. We find bringing the discriminator score towards this center value allows the optimization to find values of $\sigma$ that balance realism and variation. 
Thus, we set the objective to minimize the mean-squared L2 norm of the discriminator score vector. For a set of $k$ original latent vectors $\mathbf{Z}_{0} \in \mathbb{R}^{k \times n}$ and the protoimage vectors $\mathbf{Z}_{p} \in \mathbb{R}^{k \times n}$ where each row is a vector, we can define the optimization problem as
\begin{equation}
    \mathbf{\sigma}^{*} = \arg \min_{\mathbf{\sigma}\in [0,1]^{n}} \frac{1}{k} \sum^{k}_{i=0} D(G(\sigma_{i} \mathbf{Z}_{p,i} + (1-\sigma_{i}) \mathbf{Z}_{0,i}))^{2}
\label{eq:Optimization for best image}
\end{equation}

where $\mathbf{ \sigma } \in [0,1]^{k}$ is a vector of linear interpolation coefficients with elements $\sigma_{i}$, $\mathbf{Z}_{p,i}$ is the $i^{th}$ row of  $\mathbf{Z}_{p}$, and $\mathbf{Z}_{0,i}$ is the $i^{th}$ row of  $\mathbf{Z}_{0}$. Results of the optimization can be seen in Fig. \ref{fig:Optimized Images} along with the resulting values of $\sigma$. It can be seen that images that are already high-quality have smaller values of $\sigma$ while poor images receive larger values. 

\section{Experimental Results}

We apply our framework to a multitude of experiments to evaluate the effectiveness of the image quality improvements, the possible use cases, and the limitations of our method. 

For the latent vector recovery, we use an Adam optimizer \cite{DBLP:journals/corr/KingmaB14} with learning rate of $0.01$, $\beta_{1}=0.9$, and $\beta_{2}=0.999$. The $\alpha$ and $\beta$ coefficients from equation \ref{eq:ObjectiveFuncRecovery} are set to 2 and 1, respectively. For the initialization, we randomly generate $1,000$ latent vectors and chose the latent vector with the smallest Euclidean distance to the original image in the feature space of the discriminator. In searching for the \x images, we use an Adam optimizer with a learning rate of $0.1$, $\beta_{1}=0.9$, $\beta_{2}=0.999$, and a $\lambda$ of 3. This algorithm must be applied to a batch of latent vectors in order for the minibatch standard deviation layer to give the \x images small discriminator scores. Finally, for finding the linear interpolation coefficient, we use a learning rate of $0.01$ and set the initial values of $\sigma$ to 0.7, again with an Adam optimizer of the same $\beta_{1}$ and $\beta_{2}$. 

\subsection{Quantitative Evaluation}
\begin{figure}[t]
\begin{center}
   \includegraphics[width=1.0\linewidth]{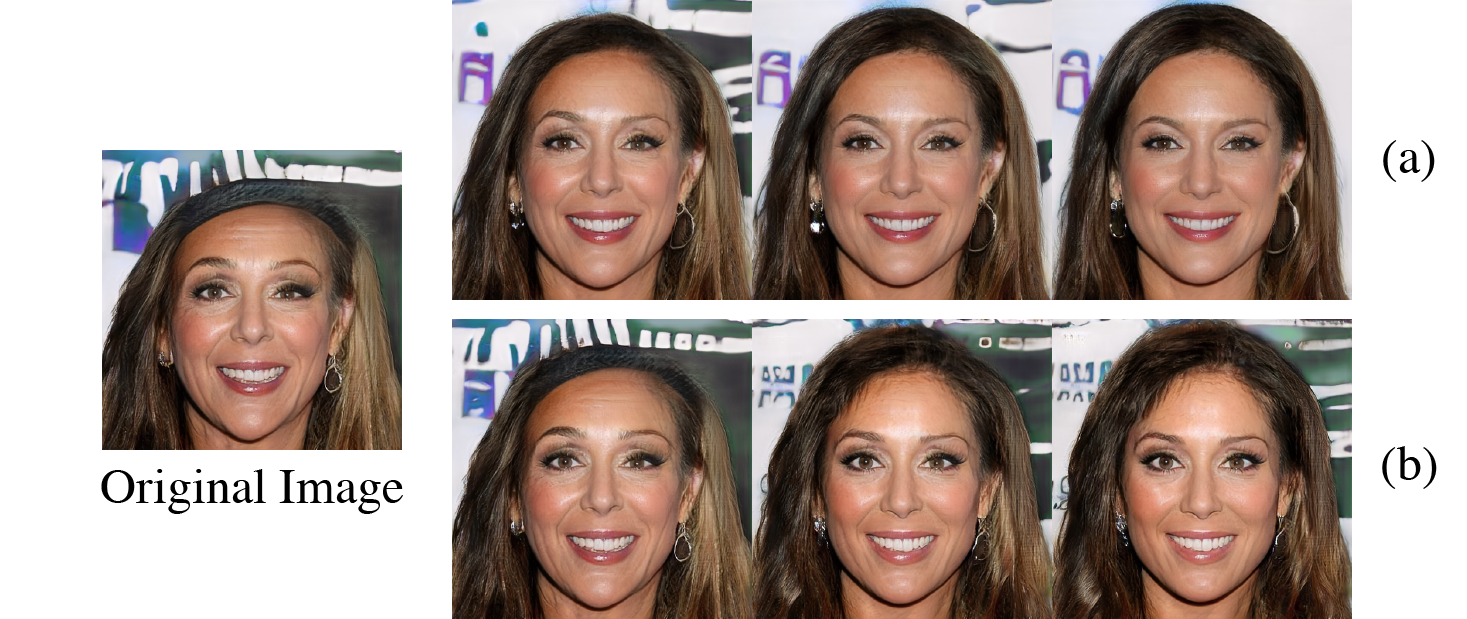}
\end{center}
   \caption{A qualitative comparison to a supervised method with CelebaHQ1024x1024 \cite{PGAN}. (a) The progression of images in the direction found by \cite{shen2020interpreting}. (b) The progression along our prescribed direction. We illustrate both a differentiation from \cite{shen2020interpreting} as well as comparable visual performance even without supervision.}
\label{fig:CelebaHQ}
\end{figure}

To evaluate the improvement of image quality and balance of image variation, we compare the Frechet Inception Distance (FID) score \cite{FID} of several variations. In Table \ref{table:fid_is}, we compare two popular GANs with and without our proposed optimization with three datasets. For the truncation trick, we take the best score given for a sequence of threshold values from 0.1 to 1.0. Similarly, for our method, we take the best score given for a series of $\sigma$ values, which gives the top score along the path towards the protoimages. We also compare our method to the top score along the direction found using the supervised approach of \cite{shen2020interpreting} on the CelebAHQ1024x1024 \cite{PGAN} dataset. Our algorithm successfully improves the FID score (by making it smaller) of the original GAN without making any changes to the network itself, and we outperform the truncation trick for all datasets and networks. Surprisingly, we are able to improve upon the performance of \cite{shen2020interpreting} even though our method is completely unsupervised. Qualitative comparisons can be seen in Fig. \ref{fig:CelebaHQ}. This result shows that the proposed algorithm consistently improves FID across different GANs and datasets.


\begin{table}
\begin{center}
\begin{tabular}{| l c c |}
\hline
Method & Dataset & FID$\downarrow$\\
\hline\hline
 PGAN & CelebA & 12.09\\ 
 \hline
 PGAN + truncation trick & CelebA & 11.84\\ 
 \hline
 PGAN + proposed method & CelebA & \textbf{11.69}\\
 \hline
 \hline
 PGAN  & Church & 10.70\\
 \hline
 PGAN + truncation trick & Church & 10.65\\
 \hline
 PGAN + proposed method & Church & \textbf{10.56}\\
 \hline
 \hline
 WGAN-GP & CelebA & 58.53\\  
 \hline
 WGAN-GP + truncation trick & CelebA & 58.36\\ 
 \hline
 WGAN-GP + proposed method & CelebA & \textbf{57.38} \\ 
 \hline
 \hline
 PGAN & CelebAHQ & \text{8.30} \\
 \hline
 PGAN + \cite{shen2020interpreting} & CelebAHQ & \text{8.21}\\
 \hline
 PGAN + proposed method & CelebAHQ & \textbf{8.20}\\
 \hline

\end{tabular}
\end{center}
\caption{FID scores of GANs with and without our proposed algorithm. Numbers in bold show the best scores for a given pair of GAN and dataset. The proposed algorithm consistently improves FID metric across different GANs and datasets.}
\label{table:fid_is}
\end{table}


\subsection{Attribute Control}

We explore if we can apply our framework to retrieve higher-quality images with a particular attribute using one of the methods described in Sec. \ref{section: ganmanipulation}. Following the same procedure as \cite{radford}, we find a set of generated images with a particular attribute and a set without the attribute. By taking the difference vector between the mean vectors of both sets, we can obtain a vector which adds or subtracts the attribute from a given image. However, the method in \cite{radford} does not necessarily result in realistic images. Thankfully, by applying our method on the resulting images, we can drastically increase the image realism while preserving the target attribute. 

\begin{figure}[t]
\begin{center}
   \includegraphics[width=1.0\linewidth]{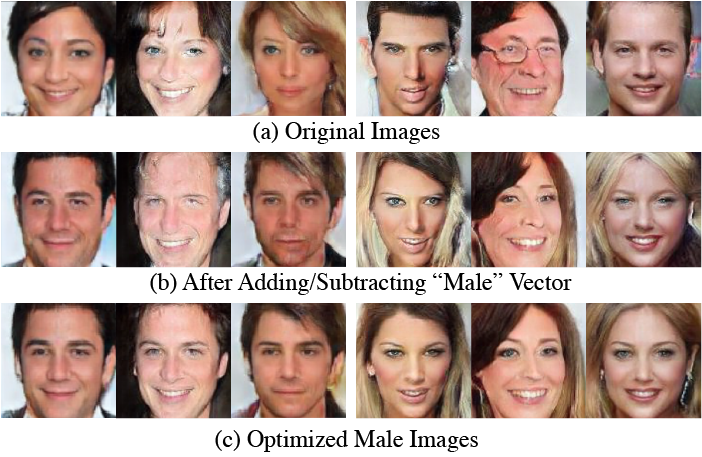}
\end{center}
   \caption{Controlling image attributes. We show the compatibility of our proposed method with an attribute control method \cite{radford}. By first adding or subtracting the ``male" vector to a randomly generated image and then applying our method, we are able to obtain high-fidelity images of a particular gender.}
\label{fig:Male Improved Images}
\end{figure}

To demonstrate this, we found a set of 100 generated images of males and 100 generated images of females. By adding the difference vector between the set averages, we can make any randomly generated image into a male. By subtracting the vector, the resulting image is of a female. Then, we apply our method to improve the photo-realism. Figs. \ref{fig:Male Improved Images} shows qualitative results of our experiment, compared to Figs. \ref{fig:Male Improved Images}(b) and \ref{fig:Male Improved Images}(c). We can see a significantly improvement in image realism with target female/male attribute preserved when using our proposed algorithm. 

\subsection{LSUN Church Dataset}
To evaluate the scope of our framework, we investigate the structure of PGAN trained on another dataset, the LSUN Church Dataset \cite{LSUN}. As the structure of the church set is much more complex than the CelebA set, we found the latent vector recovery to be much more sensitive to the initialization of the latent vector than in the CelebA set. To provide more consistent performance, we trained and utilized an encoder using the procedure of \cite{encoder2}. The encoder is able to provide a very good initial vector, and our optimization method brings the recovered image to nearly the same as the original.

\begin{figure}[t]
\begin{center}
   \includegraphics[width=1.0\linewidth]{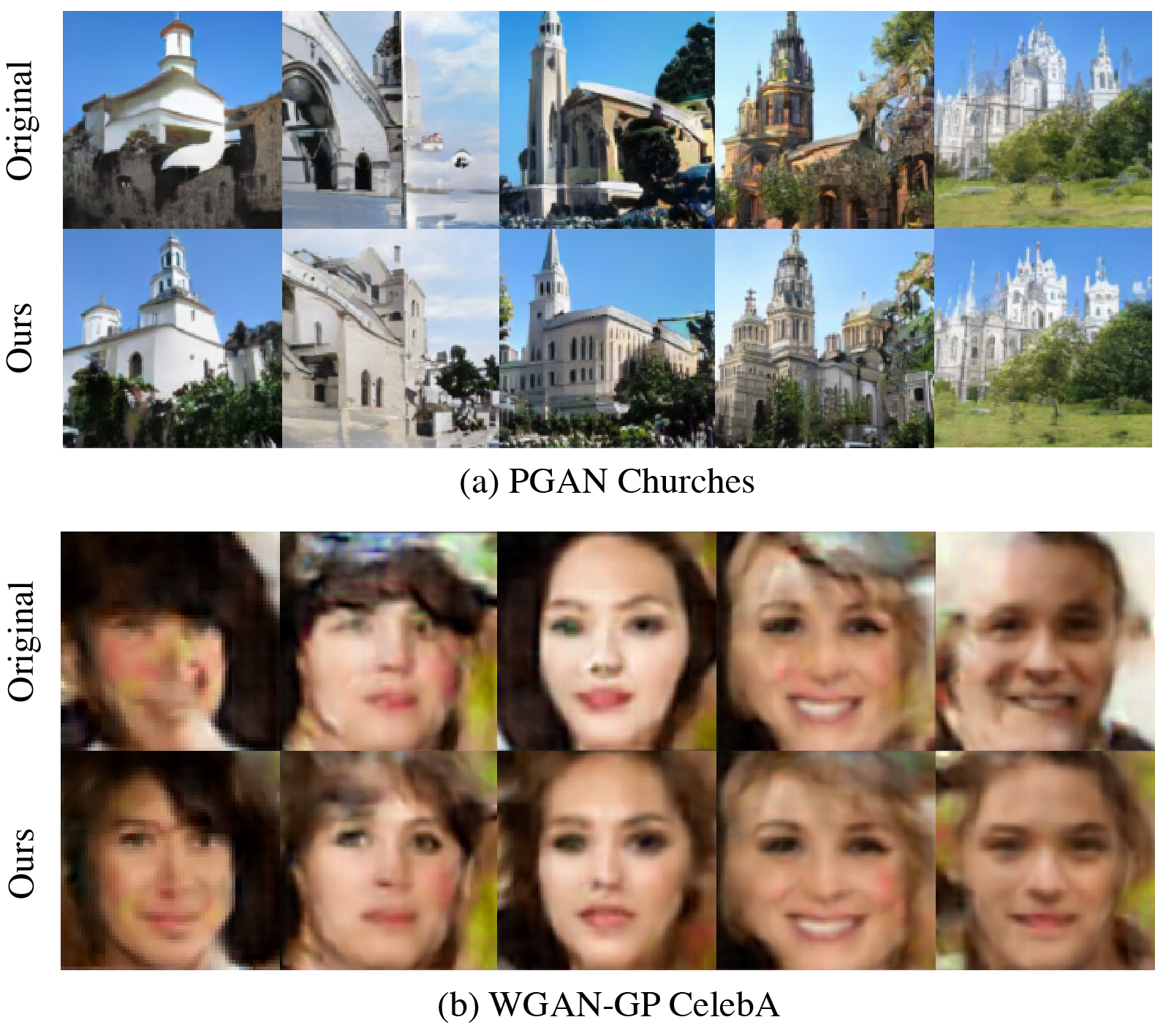}
\end{center}
   \caption{Results from multiple datasets and network architectures, showing the original output of the specified network (top rows) and the corresponding output of our method (bottom rows). Our framework is capable of improving the realism of different types of images including churches and shows compatibility with lower-complexity networks such as WGAN-GP}
\label{fig:ManyExamples}
\end{figure}

As with the CelebA set, we see a similar structure in the latent space with the presence of \x images. Following the same framework, we are able to find images of improved realism, which maintain most of the features of the original images. The outcome of this is shown in  Fig. \ref{fig:ManyExamples}. 

\subsection{Other Architectures}
We further explore the scope of our proposed algorithm as it applies to other GAN architectures. We start with one of the simplest convolution GAN,  DCGAN \cite{radford}. 

We trained a DCGAN on the CelebA dataset for 10 epochs with a 512-dimensional generator latent space to match the dimensions of the PGAN latent space. Applying our latent vector recovery algorithm and traversing the latent space as described in Section \ref{walkinlatentspace}, we identify the presence of \x images. Because the underlying space of faces is not as well estimated in DCGAN compared to PGAN, the \x images are not as distinct as those found in the PGAN latent space. The \x images are more blended, so the image quality does not necessarily improve as the latent vectors approach the \x image latent vectors.

Training with the WGAN-GP loss \cite{WGAP-GP}, the network provides more distinct \x images. As seen with the PGAN, the images improve as they approach the \x images, resembling more of a face when the original image is distorted. Results of our framework with the WGAN can be found in the Fig. \ref{fig:ManyExamples}. 

\begin{figure}[t]
\begin{center}
   \includegraphics[width=1.0\linewidth]{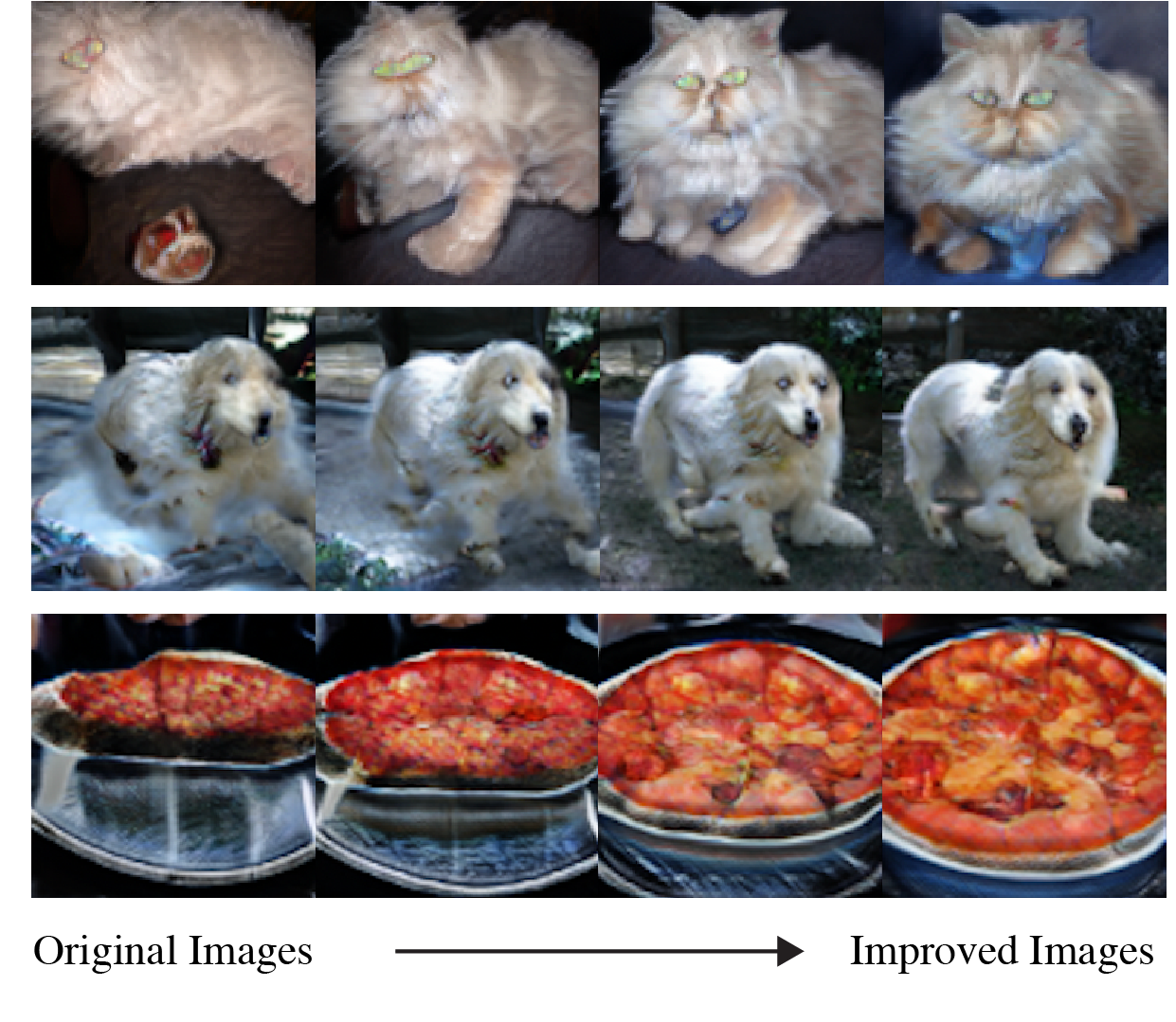}
\end{center}
\vspace{-0.8em}
   \caption{Application to class-conditioned BigGAN. Moving along the hypersphere in the direction found in Sec \ref{section: recovery} brings the image to a space that better resembles the specific class.}
\label{fig:BigGAN}
\end{figure}

While we have focused on enhancing the applicability of low-complexity networks, another branch of GAN accessibility is making the most of existing networks. Thus, we investigate the generalizability and utility of our method by applying it to BigGAN \cite{BigGAN} pretrained on ImageNet \cite{imagenet}. For the latent vector recovery procedure in Sec. \ref{section: recovery}, we leave the class-conditional code fixed when searching over the latent space.  Performing the walk as in Sec. \ref{walkinlatentspace} reveals a similar behavior as before with image realism improving until the walk reaches a protoimage. Examples can be seen in Fig. \ref{fig:BigGAN}. This illustrates the generality of the latent space direction we define. Without a minibatch standard deviation layer in the discriminator, our method can be simplified to a user-controlled manipulation as in \cite{shen2020interpreting,shen2021closedform,yang2019semantic,ganalyze,Voynov2020UnsupervisedDO} allowing users to specify the amount to travel in our discovered direction.

\section{Conclusion}

We presented a novel method for navigating the GAN latent space to increase the photo-realism of images. While previous work has achieved improvements through increasing the model complexity, our framework improves photo-realism on even shallow networks. Paired with the ability to control the attributes of generated images, we provide new insights into the structure of GAN's latent space by demonstrating the existence of \x images. By understanding the structure of the latent space, we open opportunities to manipulate the structure during training in order to provide higher-realism and improve class variability. We showed the effectiveness of our method with different architectures and different datasets. The results presented above show great improvements on realism, especially in datasets with controlled pose structure such as faces as \x images resemble the object itself.

{\small
\bibliographystyle{ieee_fullname}
\bibliography{GANsRef}

\begin{thebibliography}{10}\itemsep=-1pt

\bibitem{WGAN}
M. Arjovsky, S. Chintala, and L. Bottou.
\newblock {W}asserstein generative adversarial networks.
\newblock In {\em Proceedings of Machine Learning Research}, volume~70, pages
  214--223, 2017.

\bibitem{encoder2}
D. Bau, JY. Zhu, J. Wulff, W. Peebles, H. Strobelt, B. Zhou, and A. Torralba.
\newblock Seeing what a {GAN} cannot generate.
\newblock In {\em International Conference Computer Vision (ICCV)}, 2019.

\bibitem{bishop:2006:PRML}
Christopher~M. Bishop.
\newblock {\em Pattern Recognition and Machine Learning}, pages 36--38.
\newblock Springer, 2006.

\bibitem{BigGAN}
A. Brock, J. Donahue, and K. Simonyan.
\newblock Large scale {GAN} training for high fidelity natural image synthesis.
\newblock In {\em International Conference on Learning Representations (ICLR)},
  2019.

\bibitem{Inversion2}
A. {Creswell} and A.~A. {Bharath}.
\newblock Inverting the generator of a generative adversarial network.
\newblock {\em IEEE Transactions on Neural Networks and Learning Systems},
  30(7):1967--1974, 2019.

\bibitem{imagenet}
J. Deng, W. Dong, R. Socher, L.-J. Li, K. Li, and L. Fei-Fei.
\newblock {ImageNet: A Large-Scale Hierarchical Image Database}.
\newblock In {\em CVPR09}, 2009.

\bibitem{Donahue:2019a}
J. Donahue and K. Simonyan.
\newblock Large scale adversarial representation learning.
\newblock In {\em Advances in Neural Information Processing Systems (NeurIPS)},
  2019.

\bibitem{Dosovitskiy:2016a}
A. Dosovitskiy and T. Brox.
\newblock Generating images with perceptual similarity metrics based on deep
  networks.
\newblock In {\em Advances in Neural Information Processing Systems (NIPS)},
  2016.

\bibitem{ganalyze}
Lore Goetschalckx, Alex Andonian, Aude Oliva, and Phillip Isola.
\newblock Ganalyze: Toward visual definitions of cognitive image properties.
\newblock {\em arXiv preprint arXiv:1906.10112}, 2019.

\bibitem{goodfellow}
I.~J. Goodfellow, J. Pouget-Abadie, M. Mirza, B. Xu, D. Warde-Farley, S. Ozair,
  A. Courville, and Y. Bengio.
\newblock Generative adversarial {Nets}.
\newblock In {\em International Conference on Neural Information Processing
  Systems (NIPS)}, 2014.

\bibitem{WGAP-GP}
I. Gulrajani, F. Ahmed, M. Arjovsky, V. Dumoulin, and AC. Courville.
\newblock Improved training of wasserstein {GANs}.
\newblock In {\em Advances in neural information processing systems (NIPS)},
  2017.

\bibitem{FID}
M. Heusel, H. Ramsauer, T. Unterthiner, B. Nessler, and S. Hochreiter.
\newblock {GANs} trained by a two time-scale update rule converge to a local
  nash equilibrium.
\newblock In {\em International Conference on Neural Information Processing
  Systems (NIPS)}, 2017.

\bibitem{harkonen2020ganspace}
Erik Härkönen, Aaron Hertzmann, Jaakko Lehtinen, and Sylvain Paris.
\newblock Ganspace: Discovering interpretable gan controls.
\newblock In {\em Proc. NeurIPS}, 2020.

\bibitem{gansteerability}
Ali Jahanian, Lucy Chai, and Phillip Isola.
\newblock On the "steerability" of generative adversarial networks.
\newblock In {\em International Conference on Learning Representations}, 2020.

\bibitem{Johnson:2016}
J. Johnson, A. Alahi, and L. Fei-Fei.
\newblock Perceptual losses for real-time style transfer and super-resolution.
\newblock In {\em European Conference on Computer Vision (ECCV)}, 2016.

\bibitem{Karnewar:2020a}
A. Karnewar, O. Wang, and R. Iyengar.
\newblock {MSG-GAN}: Multi-scale gradient gan for stable image synthesis.
\newblock In {\em IEEE/CVF Conference on Computer Vision and Pattern
  Recognition}, 2020.

\bibitem{PGAN}
T. Karras, T. Aila, S. Laine, and J. Lehtinen.
\newblock Progressive growing of {GANs} for improved quality, stability, and
  variation.
\newblock In {\em International Conference on Learning Representations (ICLR)},
  2018.

\bibitem{KarrasLAHLA19}
T. {Karras}, S. {Laine}, and T. {Aila}.
\newblock A style-based generator architecture for generative adversarial
  networks.
\newblock In {\em IEEE/CVF Conference on Computer Vision and Pattern
  Recognition (CVPR)}, 2019.

\bibitem{KarrasLAHLA20}
T. Karras, S. Laine, M. Aittala, J. Hellsten, J. Lehtinen, and T. Aila.
\newblock Analyzing and improving the image quality of stylegan.
\newblock In {\em {IEEE/CVF} Conference on Computer Vision and Pattern
  Recognition (CVPR)}, 2020.

\bibitem{DBLP:journals/corr/KingmaB14}
DP. Kingma and J. Ba.
\newblock Adam: {A} method for stochastic optimization.
\newblock In {\em International Conference on Learning Representations (ICLR)},
  2015.

\bibitem{Inversion1}
Z.~C. Lipton and S. Tripathi.
\newblock Precise recovery of latent vectors from generative adversarial
  networks.
\newblock In {\em International Conference on Learning Representations
  Workshops (ICLRW)}, 2017.

\bibitem{CelebA}
Z. Liu, P. Luo, X. Wang, and X. Tang.
\newblock Deep learning face attributes in the wild.
\newblock In {\em International Conference on Computer Vision (ICCV)}, 2015.

\bibitem{Inversion3}
F. Ma, U. Ayaz, and S. Karaman.
\newblock Invertibility of convolutional generative networks from partial
  measurements.
\newblock In {\em Advances in Neural Information Processing Systems (NIPS)},
  2018.

\bibitem{pulse}
S. Menon, A. Damian, S. Hu, N. Ravi, and C. Rudin.
\newblock Pulse: Self-supervised photo upsampling via latent space exploration
  of generative models.
\newblock In {\em IEEE/CVF Conference on Computer Vision and Pattern
  Recognition (CVPR)}, 2020.

\bibitem{plumerault20iclr}
Antoine Plumerault, Herv{\'e} {Le Borgne}, and C{\'e}line Hudelot.
\newblock Controlling generative models with continuous factors of variations.
\newblock In {\em International Conference on Machine Learning (ICLR)}, 2020.

\bibitem{Qi:2019a}
G. Qi.
\newblock Loss-sensitive generative adversarial networks on lipschitz
  densities.
\newblock {\em International Journal of Computer Vision}, 128:1118--1140, 2019.

\bibitem{radford}
A. Radford, L. Metz, and S. Chintala.
\newblock Unsupervised representation learning with deep convolutional
  generative adversarial networks.
\newblock In {\em International Conference on Learning Representations (ICLR)},
  2016.

\bibitem{Razavi:2019a}
A. Razavi, A. van~den Oord, and O. Vinyals.
\newblock Generating diverse high-fidelity images with vq-vae-2.
\newblock In {\em Advances in Neural Information Processing Systems (NeurIPS)},
  2019.

\bibitem{shen2020interpreting}
Yujun Shen, Jinjin Gu, Xiaoou Tang, and Bolei Zhou.
\newblock Interpreting the latent space of gans for semantic face editing.
\newblock In {\em CVPR}, 2020.

\bibitem{shen2021closedform}
Yujun Shen and Bolei Zhou.
\newblock Closed-form factorization of latent semantics in gans.
\newblock In {\em CVPR}, 2021.

\bibitem{VGG16}
K. Simonyan and A. Zisserman.
\newblock Very deep convolutional networks for large-scale image recognition.
\newblock In {\em International Conference on Learning Representations (ICLR)},
  2015.

\bibitem{Voynov2020UnsupervisedDO}
A. Voynov and A. Babenko.
\newblock Unsupervised discovery of interpretable directions in the gan latent
  space.
\newblock {\em ArXiv}, abs/2002.03754, 2020.

\bibitem{yang2019semantic}
Ceyuan Yang, Yujun Shen, and Bolei Zhou.
\newblock Semantic hierarchy emerges in deep generative representations for
  scene synthesis.
\newblock {\em International Journal of Computer Vision}, 2020.

\bibitem{LSUN}
F. Yu, Y. Zhang, S. Song, A. Seff, and J. Xiao.
\newblock Lsun: Construction of a large-scale image dataset using deep learning
  with humans in the loop.
\newblock {\em arXiv preprint arXiv:1506.03365}, 2015.

\bibitem{encoder1}
JY Zhu, P. Kr{\"a}henb{\"u}hl, E. Shechtman, and A.~A. Efros.
\newblock Generative visual manipulation on the natural image manifold.
\newblock In {\em European Conference on Computer Vision (ECCV)}, 2016.

\bibitem{zhu2020indomain}
Jiapeng Zhu, Yujun Shen, Deli Zhao, and Bolei Zhou.
\newblock In-domain gan inversion for real image editing.
\newblock In {\em Proceedings of European Conference on Computer Vision
  (ECCV)}, 2020.

\end{thebibliography}
}

\clearpage
\setcounter{section}{0}
\setcounter{equation}{0}
\setcounter{figure}{0}
\setcounter{equation}{0}

\renewcommand{\thetable}{S\arabic{table}}
\renewcommand{\thefigure}{S\arabic{figure}}
\renewcommand{\theequation}{S\arabic{equation}}
\renewcommand{\thesection}{S\arabic{section}}

\section{Summary}
In Section \ref{section: proof}, we prove that a high-dimensional Gaussian distribution is a soap-bubble. Section \ref{section:protoimage} discusses the traits of the \x images and show examples of the \x images for different datasets. Section \ref{section:implementation} provides implementation details of our method with different architectures and datasets. Section \ref{section: variations} describes variations of our method that can improve speed and user control. Finally, Section \ref{section:figures} includes additional images of our results with more side-by-side comparisons of randomly generated images and our optimized output.

\begin{figure*}[h!]
\begin{center}
   \includegraphics[width=1.0\linewidth]{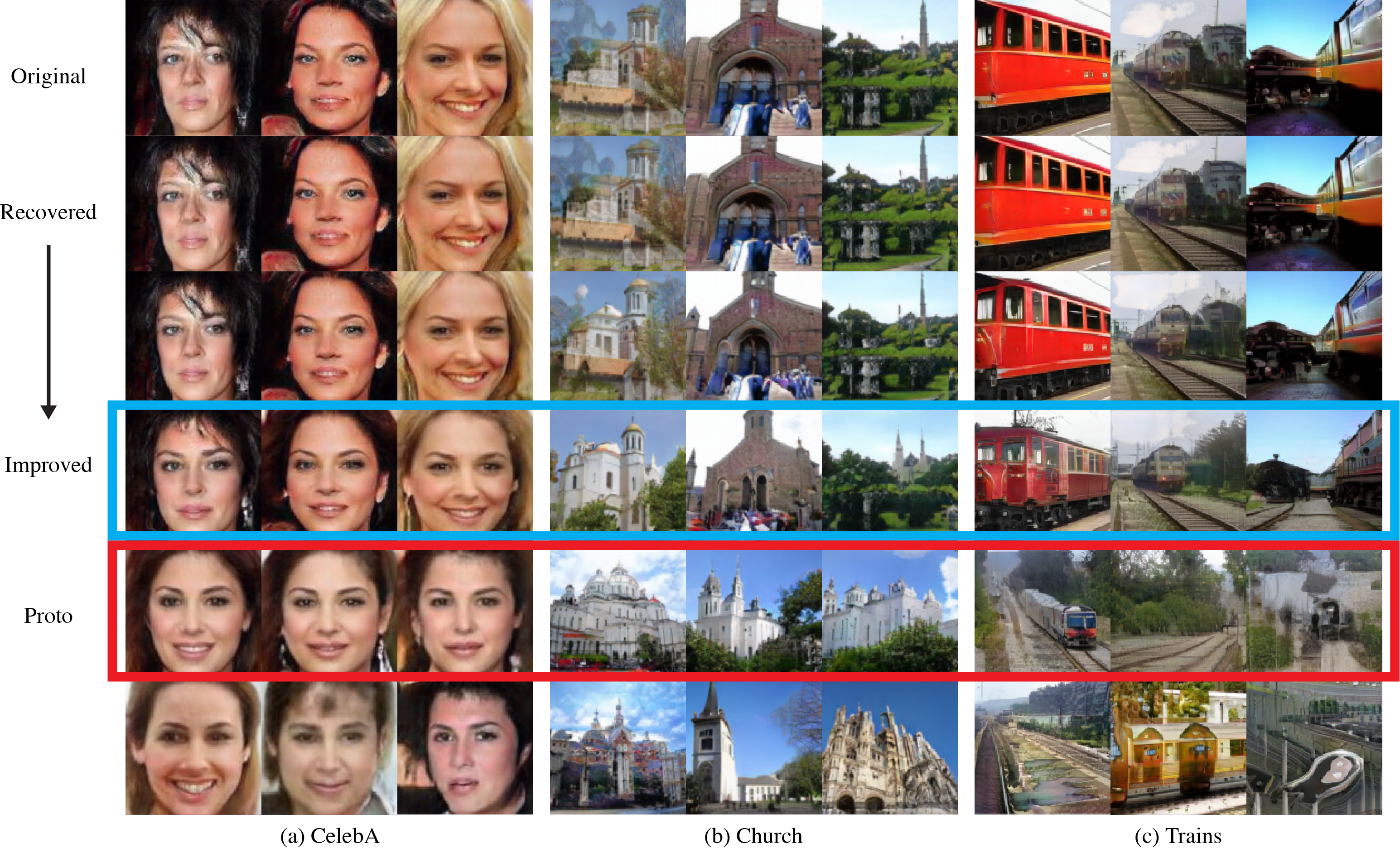}
\end{center}
   \caption{Here we show the progression as we traverse the latent space. The image quality improves as we approach the \x images marked in red. Note the similarity of the \x images. We see the step after the \x images shows a completely different identity or scene.}
\label{fig:TraverseImages}
\end{figure*}

\section{Proof for High-dimensional Gaussians} \label{section: proof}
As previously mentioned, the majority of the probability mass of high-dimensional Gaussians lie within a thin shell of a hypersphere. Due to this phenomena, our method of traversing the latent space becomes a simple walk along the great circles of a hypersphere. Here, we provide a formal proof of this mathematical fact. Let $\mathbf{X}=[X_1, ..., X_n]$ where $X_i \sim N(0,1)$ are i.i.d. Denote the norm of this vector as 
\begin{equation}
    Q = \|\mathbf{X}\|_2 = \sqrt{X_{1}^{2} + ... + X_{n}^{2}}
\end{equation} 
$Q$ follows a chi distribution with variance 
\begin{equation}
    \text{var}(Q) = n - E[Q]^2
\end{equation}
where $E[\cdot]$ denotes the expected value.
The variance for a chi distribution tends toward $\frac{1}{2}$ for large $n$, which gives 
\begin{equation}
    E(Q) \approx \sqrt{n-\frac{1}{2}} \approx \sqrt{n}
\end{equation}
Thus, the norm of $\mathbf{X}$ follows a distribution with a mean of approximatedly $\sqrt{n}$ and a very small variance, proving that $\mathbf{X}$ must have the majority of its probability mass in a thin shell around a hypersphere of radius $\sqrt{n}$. Additional and alternate proofs can be found in \cite{bishop:2006:PRML,shen2020interpreting,plumerault20iclr}. 

\section{The Protoimage} \label{section:protoimage}
Here we present a more detailed discussion and analysis of the \x images. These images serve as a step in our method of improving image realism, but their existence and qualities are interesting in themselves, warranting further study. 

\begin{figure}
\begin{center}
   \includegraphics[width=1.0\linewidth]{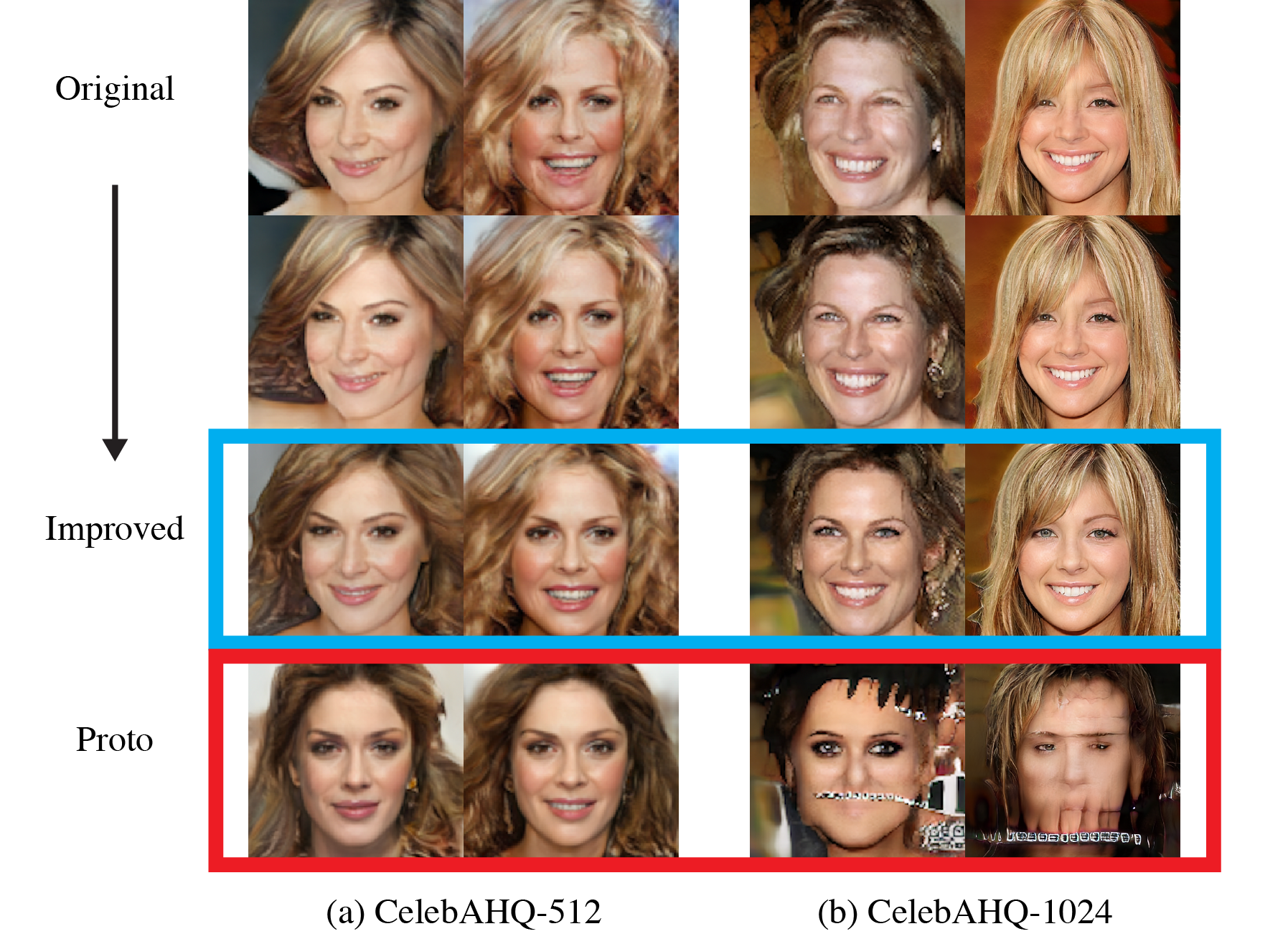}
\end{center}
   \caption{We observe a similar traversal progression with larger resolution datasets such as CelebAHQ512 and CelebAHQ1024. The 1024x1024 dataset presents unique artifacts to the protoimages; however, the preceeding images still improve in realism.}
\label{fig:CelebAHQTraverse}
\end{figure}

By following our procedure for traversing the latent space in Section 3.2 of the main paper, we are able to observe the \x image along the path around the hypersphere. This effect can be seen in Fig. \ref{fig:TraverseImages}. The \x images are very similar in appearance and seem to represent a sample mean. Despite the similarity between these images, the corresponding latent vectors are distinct and a large distance from one another.  As we approach the \x images, we can observe an improvement in image fidelity. The \x images are observed in the same iteration of the traversal, and the proceeding iteration yields a new identity or scene, as shown in the last row of Fig. \ref{fig:TraverseImages}. Additionally, we observe that the \x images take the attributes of the majority class within the training set. For the CelebA dataset, the \x images are all female as the dataset contains more images of female celebrities than male. Likewise,  the \x images of the LSUN Church dataset are images of white churches as the majority of churches in the dataset are white. Interestingly, we observe changes in the \x images for PGAN trained on different resolutions of CelebAHQ. For CelebAHQ 512x512, the \x exhibit the same behavior as previously described. For CelebAHQ 1024x1024, the \x images start to include a unique artifact, Fig. \ref{fig:CelebAHQTraverse}. Despite the presence of the artifact, we do find that moving towards the \x image still improves the realism of the face but warrants earlier stopping along the path.

\section{Implementation Details} \label{section:implementation}
We outline the different hyperparameters used for different networks and datasets. For the PGAN with CelebA, we used $\alpha =2$, $\beta = 1$, $\lambda = 3$. For the PGAN with LSUN Church and Train, we used the same values for $\alpha$ and $\beta$ but set $\lambda = 5$. For the PGAN with CelebAHQ, we used $\alpha = 10$, $\beta = 1$, and $\lambda = 100$. While using BigGAN pretrained on ImageNet, the hyperparameters were $\alpha = 15$, $\beta = 1$, and $\lambda = 100$. When searching for the protoimage with WGAN-GP trained on CelebA, we found we needed to add two addition terms to the loss function to prevent overshooting the protoimage. For this case, we add a term for the standard deviation of the discriminator scores and standard deviation of the cosine similarity between the original latent vectors and the current latent vectors in the batch. This allows the latent vectors to move along the hypersphere at a similar pace, which makes the detection of the protoimages more apparent. For a set of $k$ original latent vectors $\mathbf{Z}_{0} \in \mathbb{R}^{k \times n}$ and the protoimage vectors $\mathbf{Z}_{p}^{*} \in \mathbb{R}^{k \times n}$ where each row is a vector, the optimization can be refined as
\begin{equation}
\begin{aligned}
    \mathbf{Z}_{p}^{*} = \arg \min_{\mathbf{Z}_{p}} \frac{1}{k} \sum^{k}_{i=0} D(G(\mathbf{Z}_{p,i})) + \lambda \frac{1}{k} \sum^{k}_{i=0} M(\mathbf{Z}_{p,i},\mathbf{Z}_{0,i}) \\ + \gamma\, \text{std}(D(G(\mathbf{Z}_{p})) + \delta\,\text{std}(M(\mathbf{Z}_{p},\mathbf{Z}_{0}))
\end{aligned}
\label{eq:Optimization for best image}
\end{equation}
where $\gamma$ and $\delta$ are hyperparameters, $M(\cdot,\cdot)$ is the pairwise cosine similarity as before, and $\text{std}(\cdot)$ is the standard deviation. We set $\lambda = 0.2$, $\gamma = 1$, and $\delta = 3$.

We used the PGAN and WGAN-GP implementations from \url{https://github.com/facebookresearch/pytorch_GAN_zoo.git}. The model weights for CelebA, CelebAHQ256, and CelebAHQ512 were taken from the same repository. The weights for CelebAHQ1024 and all LSUN datasets were ported from \url{https://github.com/tkarras/progressive_growing_of_gans.git}. The weights were converted from Tensorflow to PyTorch. For BigGAN, we used the author's PyTorch implementation at \url{https://github.com/ajbrock/BigGAN-PyTorch.git}. All networks and our method were capable of running on a single NVIDIA RTX 2070 graphics card. 

\section{Method Variations} \label{section: variations}
In Section 3.4 of the main paper, our method finds the latent vector of the protoimage for each starting sample. While this is effective, we alternatively find that the procedure can be simplified for applications to a large number of samples. Instead, we can find the protoimage latent vectors for a subset of 1000 images and calculate the mean latent vector. Then, we can apply the procedures in Section 3.2 or 3.4 to move the samples towards the mean latent vector. This results in a similar improvement in photo-realism while reducing the required computational load. 

The optimization in Section 3.4 can be further controlled by the user by defining bounds on the value of $\sigma$. If the user wants to guarantee some level of change in the image without allowing the image to get too close to the protoimage, they can place an lower and upper bound on the value of $\sigma$. During the optimization, $\sigma$ would simply be projected back into the set if it were to exceed the bounds. We find this variation to be a balance between full user control as in \cite{shen2020interpreting,shen2021closedform,yang2019semantic,ganalyze,Voynov2020UnsupervisedDO} and our automated method.

\section{Additional Figures} \label{section:figures}
\begin{figure*}[t]
\begin{center}
   \includegraphics[width=0.8\linewidth]{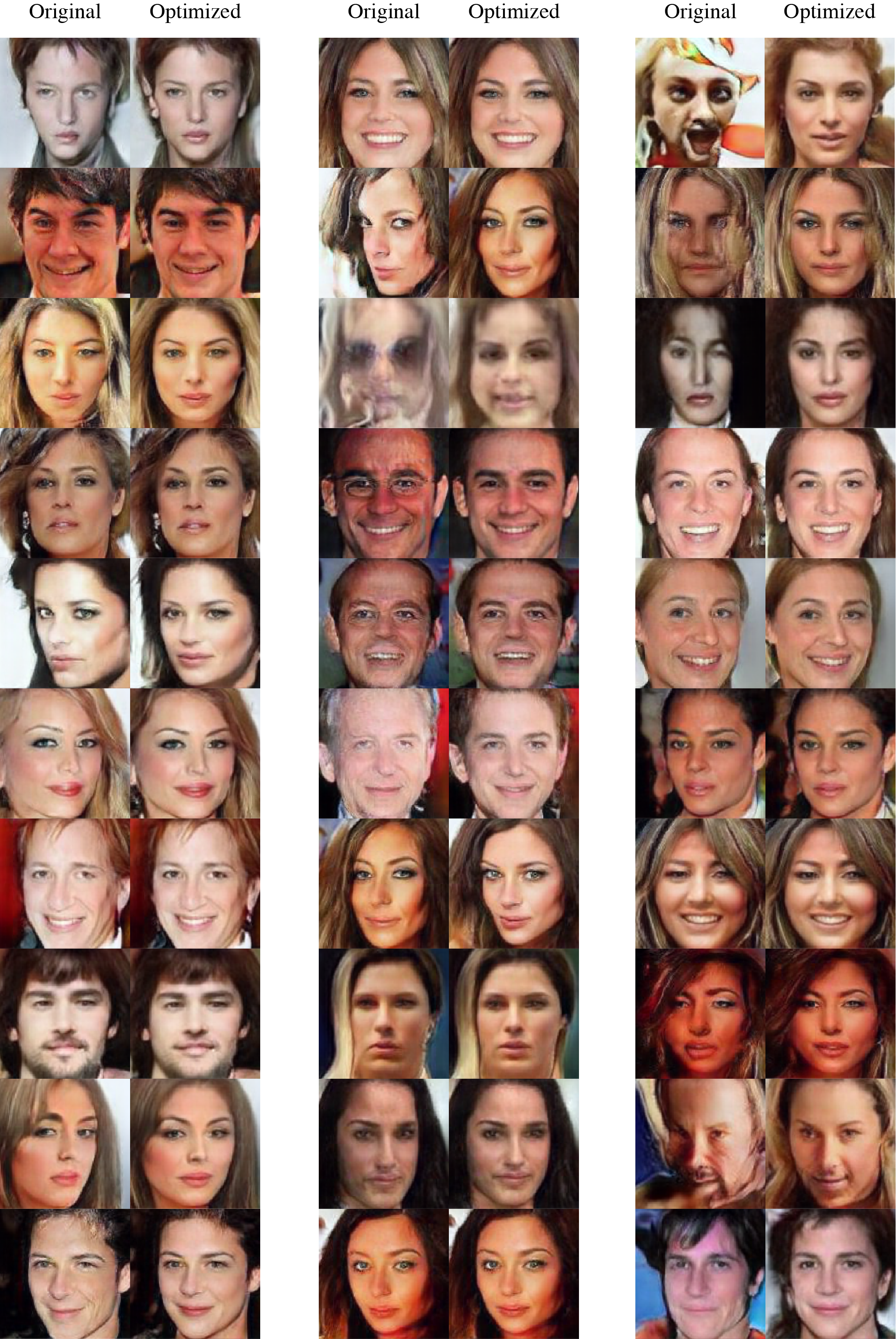}
\end{center}
   \caption{A set of 30 randomly generated images from a PGAN pretrained on CelebA with the corresponding output of our method.}
\label{fig:Uncurated}
\end{figure*}

\begin{figure*}[t]
\begin{center}
   \includegraphics[width=0.8\linewidth]{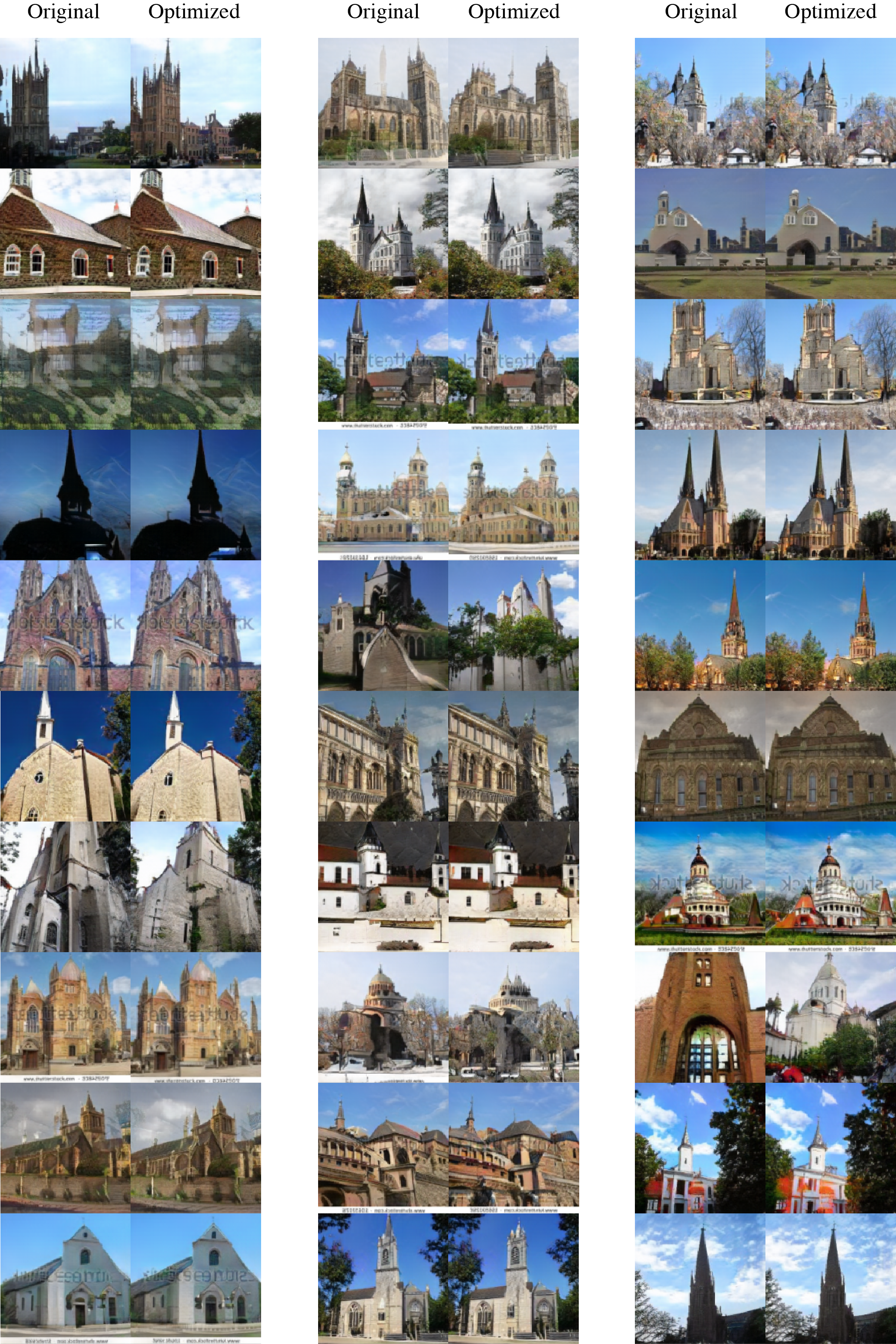}
\end{center}
   \caption{A set of 30 randomly generated images from a PGAN pretrained on LSUN Church with the corresponding output of our method.}
\label{fig:Uncurated Church}
\end{figure*}

\begin{figure*}[t]
\begin{center}
   \includegraphics[width=0.8\linewidth]{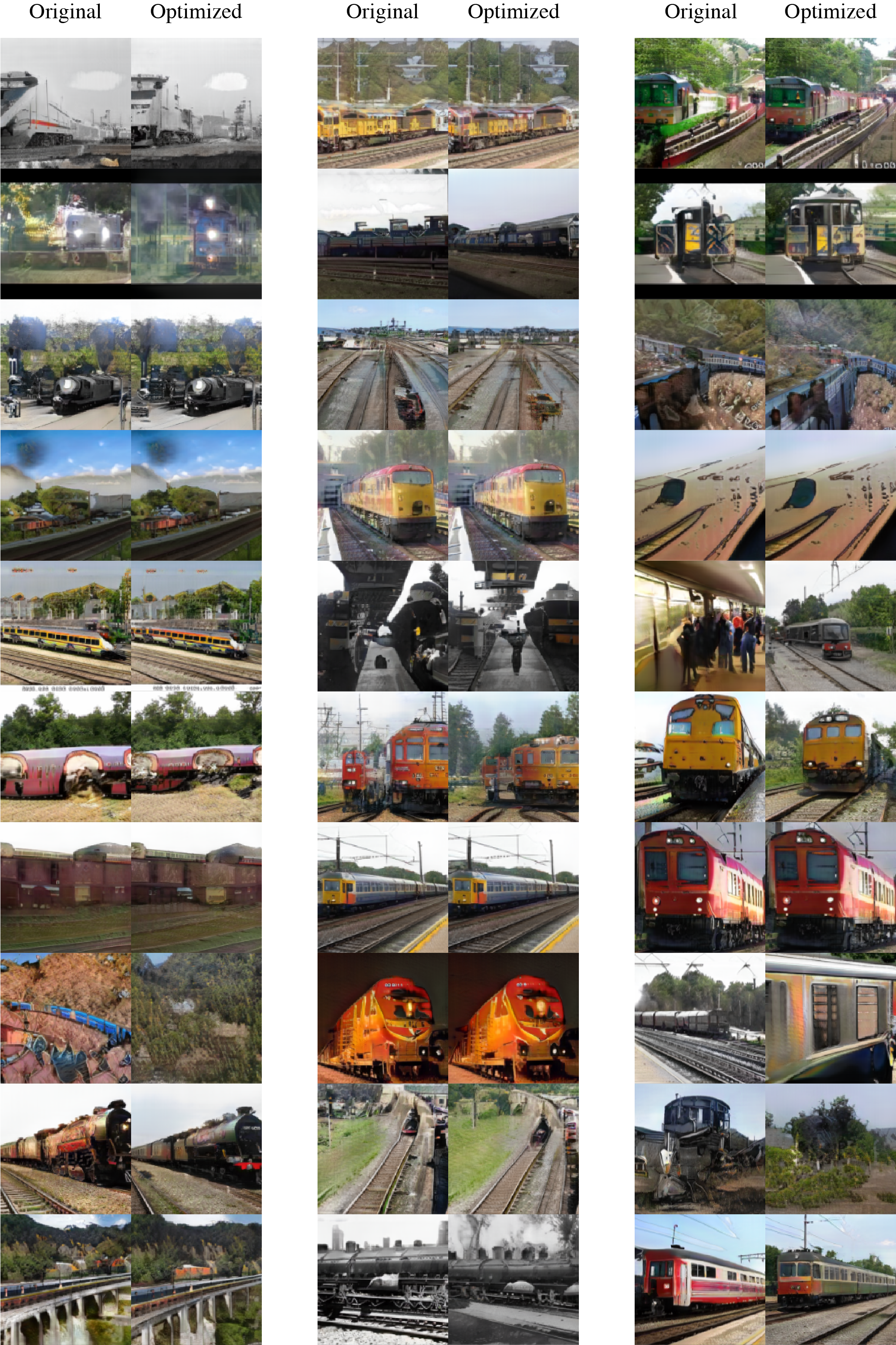}
\end{center}
   \caption{A set of 30 randomly generated images from a PGAN pretrained on LSUN Train with the corresponding output of our method.}
\label{fig:Uncurated Train}
\end{figure*}

In Fig. \ref{fig:Uncurated}, \ref{fig:Uncurated Church}, \ref{fig:Uncurated Train} we show a set of results on the CelebA, LSUN Church, and LSUN Train dataset. We randomly generate 30 images and show the resulting output from our framework. The results show a balance of improving the realism while not allowing all the images to converge to the \x images. As a result, not all of the images experience a large transformation. The framework yields better results for datasets with objects of consistent structure such as faces; however, it is still able to improve images of datasets with more variety as seen with the churches and trains.

\end{document}